\documentclass[letterpaper]{article} 
\usepackage{aaai24}  
\usepackage{times}  
\usepackage{helvet}  
\usepackage{courier}  
\usepackage[hyphens]{url}  
\usepackage{graphicx} 
\usepackage{natbib}  
\usepackage{caption} 
\usepackage{algorithm}
\usepackage{algorithmic}
\usepackage{newfloat}
\usepackage{listings}
\usepackage[export]{adjustbox} 
\usepackage{xspace}
\usepackage{graphicx}
\usepackage{amsmath}
\usepackage{amssymb}
\usepackage{booktabs}
\usepackage[dvipsnames]{xcolor}
\usepackage{xcolor,colortbl}
\usepackage{color, colortbl}
\usepackage{xcolor}
\usepackage{sidecap}

\usepackage{amsmath}
\usepackage{amssymb}
\usepackage{float}

\definecolor{darkred}{RGB}{198, 129, 129}

\definecolor{tabhighlight}{HTML}{e5e5e5}

\newcommand{\rotbox}[1]{\rotatebox{55}{#1}}
\usepackage{graphicx, amsmath, amssymb, caption, subcaption, multirow, overpic}
\definecolor{tabhighlight}{HTML}{e5e5e5}
\definecolor{citecolor}{HTML}{0071bc}

\def\eg{\emph{e.g.}\xspace}
\def\etc{\emph{etc.}\xspace}

\DeclareCaptionStyle{ruled}{labelfont=normalfont,labelsep=colon,strut=off} 
\floatstyle{ruled}
\newfloat{listing}{tb}{lst}{}
\floatname{listing}{Listing}
\title{Concept-Guided Prompt Learning for Generalization in Vision-Language Models}
\author {
    Yi Zhang\textsuperscript{\rm 1,2},
    Ce Zhang\textsuperscript{\rm 3},
    Ke Yu\textsuperscript{\rm 2},
    Yushun Tang\textsuperscript{\rm 2},
    Zhihai He\textsuperscript{\rm 2,4}\thanks{Corresponding author.}
}
\affiliations {
    \textsuperscript{\rm 1}Harbin Institute of Technology \\
    \textsuperscript{\rm 2}Southern University of Science and Technology\\
    \textsuperscript{\rm 3}Carnegie Mellon University \\
    \textsuperscript{\rm 4}Pengcheng Laboratory\\
    {zhangyi2021@mail.sustech.edu.cn, cezhang@cs.cmu.edu}\\
    {\{yuk2020, tangys2022\}@mail.sustech.edu.cn, hezh@sustech.edu.cn}
}
\usepackage{bibentry}

\begin{document}

\maketitle

\begin{abstract}
Contrastive Language-Image Pretraining (CLIP) model has exhibited remarkable efficacy in establishing cross-modal connections between texts and images, yielding impressive performance across a broad spectrum of downstream applications through fine-tuning. However, for generalization tasks, the current fine-tuning methods for CLIP, such as CoOp and CoCoOp, demonstrate relatively low performance on some fine-grained datasets. We recognize the underlying reason is that these previous methods only projected global features into the prompt, neglecting the various visual concepts, such as colors, shapes, and sizes, which are naturally transferable across domains and play a crucial role in generalization tasks. To address this issue, in this work, we propose \textbf{C}oncept-Guided \textbf{P}rompt \textbf{L}earning (CPL) for vision-language models. Specifically, we leverage the well-learned knowledge of CLIP to create a visual concept cache to enable concept-guided prompting. In order to refine the text features, we further develop a projector that transforms multi-level visual features into text features. 
We observe that this concept-guided prompt learning approach is able to achieve enhanced consistency between visual and linguistic modalities. 
Extensive experimental results demonstrate that our CPL method significantly improves generalization capabilities compared to the current state-of-the-art methods.
\end{abstract}

\section{Introduction}
Recent studies in pre-trained Vision-Language Models (VLMs), such as CLIP~\cite{radford2021learning} and ALIGN~\cite{jia2021scaling}, highlight a promising direction for foundation models in performing a variety of open-vocabulary tasks. By understanding various visual concepts learned from extensive image-text pairs, these models exhibit impressive capabilities across a broad spectrum of downstream tasks in a zero/few-shot manner~\cite{radford2021learning,alayrac2022flamingo,yu2022coca}.

Although the zero-shot CLIP model demonstrates competitive performance in various visual tasks, its nature as a pre-trained model hinders its ability to generalize to unseen domains. Therefore, several works focus on fine-tuning these pre-trained VLMs for downstream tasks through designing learnable prompts derived from training instances.

\begin{figure}[H]
\begin{center}
\centerline{\includegraphics[width=\linewidth]{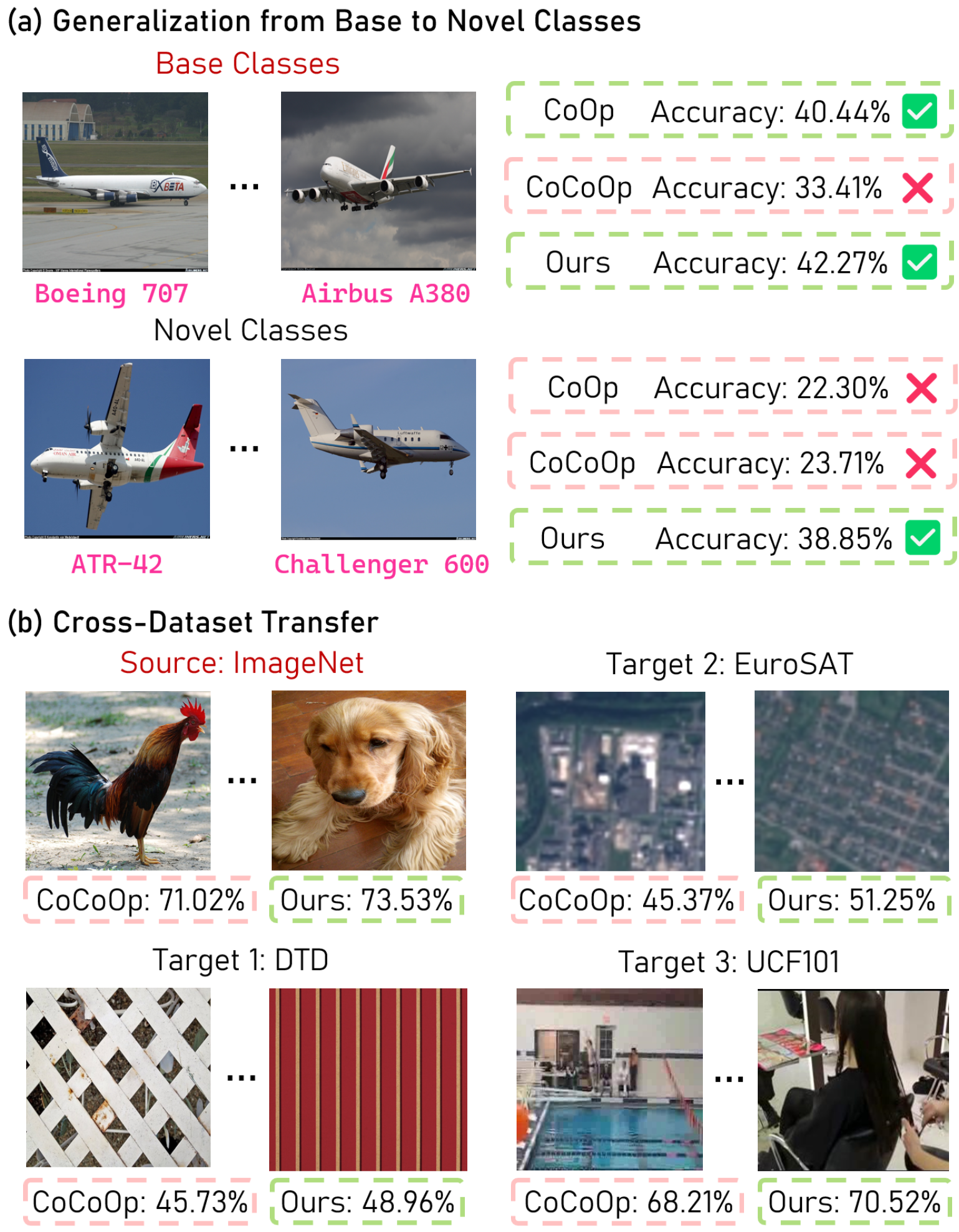}}
\caption{Examples and performance comparisons on base-to-novel generalization and cross-dataset transfer tasks. Our proposed CPL exhibits remarkable generalization capabilities in comparison to other state-of-the-art methods.}
\label{fig:comparison}
\end{center}
\end{figure}

\begin{figure*}[ht]
\begin{center}
\centerline{\includegraphics[width=\linewidth]{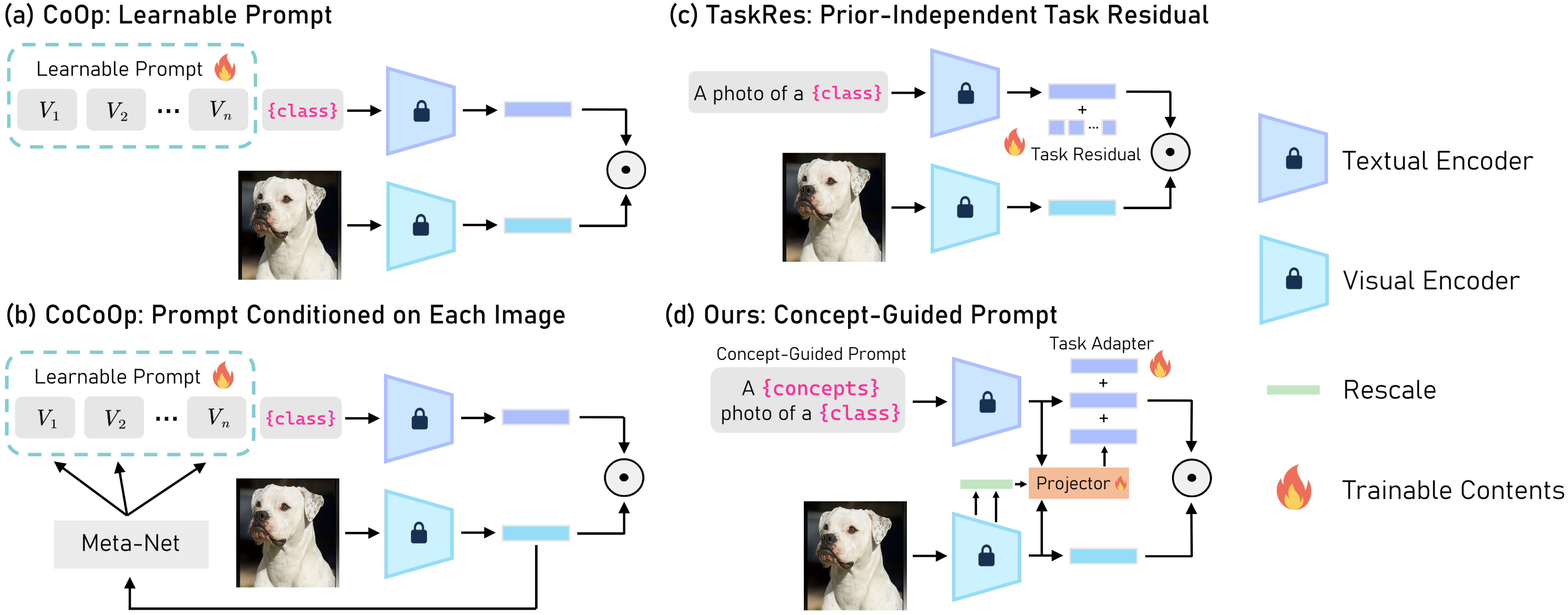}}
\caption{An illustration comparing our proposed proposed CPL approach with related baselines. We include CoOp~\cite{zhou2022learning}, CoCoOp~\cite{zhou2022conditional} and TaskRes~\cite{yu2022task} for comparison.}
\label{fig:idea}
\end{center}
\end{figure*}

\noindent 
For example, CoOp~\cite{zhou2022learning} firstly introduces learnable prompts to distill task-relevant knowledge; CoCoOp~\cite{zhou2022conditional} suggests adjusting the prompt based on each individual image; and TaskRes~\cite{yu2022task} proposes to incorporate a prior-independent task residual that doesn't undermine the well-learned knowledge of CLIP. For clarification, we provide an overview of each of the aforementioned methods in Figure \ref{fig:idea}.

For the generalization tasks as shown in Figure \ref{fig:comparison}, we observe that the current fine-tuning methods for CLIP, such as CoOp and CoCoOp, demonstrate relatively low performance on some difficult fine-grained datasets such as DTD (texture recognition), FGVC Aircraft (fine-grained classification), EuroSAT (satellite image recognition), and UCF101 (action recognition). 
We recognize that this issue may arise from CoOp and CoCoOp's direct tuning of the input text prompts to the text encoder, which can potentially undermine the previously well-learned knowledge of VLMs. To address this issue, TaskRes attempts to incorporate prior-independent learnable contexts to preserve this knowledge. To further explore the potential of prompt tuning methods, we ask: \textit{is it possible to utilize the prior knowledge of VLMs during the fine-tuning process without destroying it?}

We also observe that the previous fine-tuning methods only considered adapting to a specific task using supervised loss, which is not fully effective in generalizing to unseen domains. This limitation stems from the fact that they primarily consider class-specific features and overlook low-level visual concepts, such as colors, shapes, and materials. However, these low-level concepts are naturally transferable across domains and are therefore essential for enabling vision-language models to generalize. As a result, we are prompted to ask: \textit{is it possible to incorporate visual concepts into the fine-tuning process for VLMs to enhance their transfer capabilities?}

To address the problems above, in this work, we propose Concept-Guided Prompt Learning (CPL) for vision-language models. Specifically, we leverage the well-learned knowledge of CLIP to create a visual concept cache to enable concept-guided prompting. In order to refine the text features, we further develop a projector that projects multi-level visual features into text features. We observe that this concept-guided prompt learning approach is able to achieve enhanced consistency between visual and linguistic modalities, leading to improved generalization capability. We conducted extensive experiments to evaluate the proposed CPL approach on base-to-novel generalization, cross-dataset transfer, and domain generalization tasks. 
Our comprehensive empirical results demonstrate the significantly superior performance of CPL compared to existing state-of-the-art methods.

\section{Related Work}

\subsection{Vision-Language Models}
In recent years, vision-language models have attracted significant attention from researchers, emerging as a novel paradigm for performing visual tasks. Specifically, large-scale VLMs have been utilized to acquire general visual representations guided by natural language supervision~\cite{radford2021learning}. Current studies highlight that these models, pre-trained on vast image-text pairs available online, are capable of understanding both the semantics of images paired with their respective textual descriptions~\cite{radford2021learning,yu2022coca}. 
Recent studies~\cite{zhang2021vt,zhou2022learning} have showcased that with a robust comprehension of open-vocabulary concepts, VLMs are able to tackle various downstream visual tasks, including image retrieval~\cite{duan2022multi}, depth estimation~\cite{hu2023learning}, visual grounding~\cite{li2021grounded}, visual question answering~\cite{duan2022multi}.

\subsection{Fine-Tuning VLMs}
Fine-tuning is crucial in adapting VLMs to downstream tasks~\cite{duan2022multi}. 
Among various fine-tuning methods for VLMs, two primary approaches stand out: prompt tuning methods and adapter-based methods, respectively.

\paragraph{Prompt Tuning Methods.} Prompt tuning methods transform prompts into continuous vector representations for end-to-end objective function optimization, distilling task-relevant information from prior knowledge of VLMs~\cite{zhou2022conditional,zhou2022learning}. As the foundational work in this field, CoOp~\cite{zhou2022learning} optimizes the prompt context by a continuous set of learnable vectors. Further, CoCoOp~\cite{zhou2022conditional} recognizes the generalization issue not addressed by CoOp and proposes to generate prompts on each individual image. MaPLe~\cite{khattak2023maple} tunes both vision and language branches via a vision-language coupling function in order to induce cross-modal synergy.

\paragraph{Adapter-Based Methods.} Another series of works directly transforms the features extracted by encoders of CLIP to perform adaptation to downstream tasks. These methods are referred to as adapter-based methods. For example, CLIP-Adapter~\cite{gao2023clip}, one of the pioneering works, leverages an extra feature adapter to enhance traditional fine-tuning outcomes. 
Following CLIP-Adapter, Tip-Adapter~\cite{zhang2022tip} introduces a training-free approach by constructing a key-value cache model based on few-shot samples. CCLI~\cite{zhang2023cross} proposes to enable concept-level image representation to perform downstream tasks. BDC-Adapter \cite{zhang2023bdc} enhances vision-language reasoning by providing a more robust metric for measuring similarity between features. In addition, \citet{zhu2023not} proposes APE, which harnesses the prior knowledge of VLMs by a prior cache model, and explores the trilateral relationships among test images, the prior cache model, and textual representations.

\subsection{Visual Concept Learning}
Existing literature has suggested two major approaches to visual concept learning. The first approach typically uses hand-crafted semantic concept annotations (\eg, colors, textures, and fabric) for the training images~\cite{patterson2012sun,patterson2016coco,pham2021learning}, which is labor-intensive in practice. To address this issue, researchers propose the second approach, which aims at designing data-driven concepts through unsupervised learning~\cite{fei2005bayesian,liu2011recognizing,huang2016unsupervised}.  While these acquired concepts might initially appear sensible, they can often carry inherent biases, ultimately constraining their overall performance. Empowered by CLIP~\cite{radford2021learning}, in this work, we design an unsupervised concept mining-and-cache technique that is capable of discovering a large set of visual concepts with semantics corresponding to pre-defined text concepts.

\section{Method}
\begin{figure*}[ht]
\begin{center}
\centerline{\includegraphics[width=0.98\linewidth]{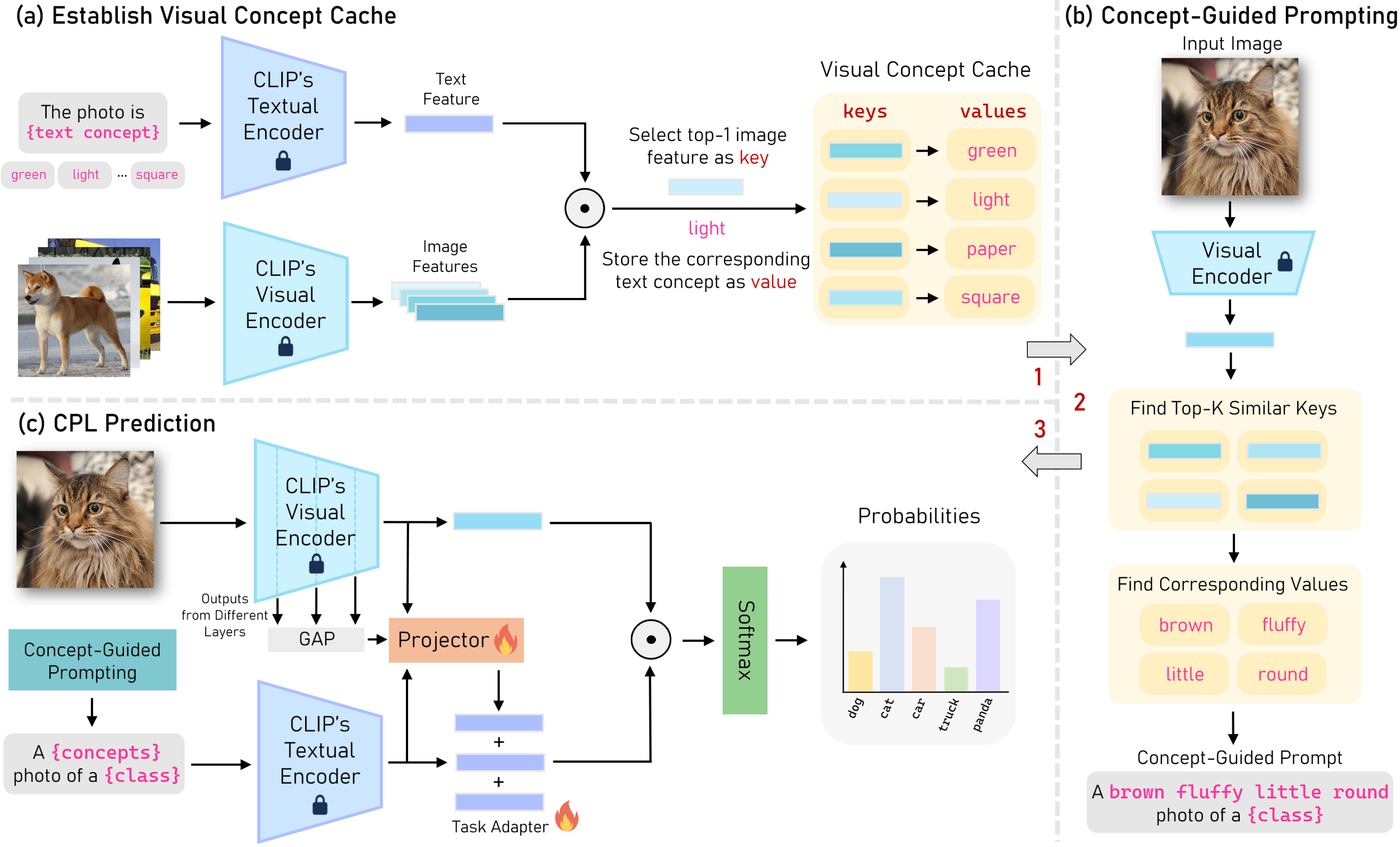}}
\caption{An overview of our proposed Concept-Guided Prompt Learning (CPL) method. Subfigure (a) shows the visual concept cache-establishing process. Subfigure (b) shows the concept-guided prompt discovery process. Subfigure (c) presents the training pipeline of our proposed CPL, where the projector and task adapter are learnable.}
\label{fig:overview}
\end{center}
\end{figure*}

\begin{figure}[ht]
\begin{center}
\centerline{\includegraphics[width=\linewidth]{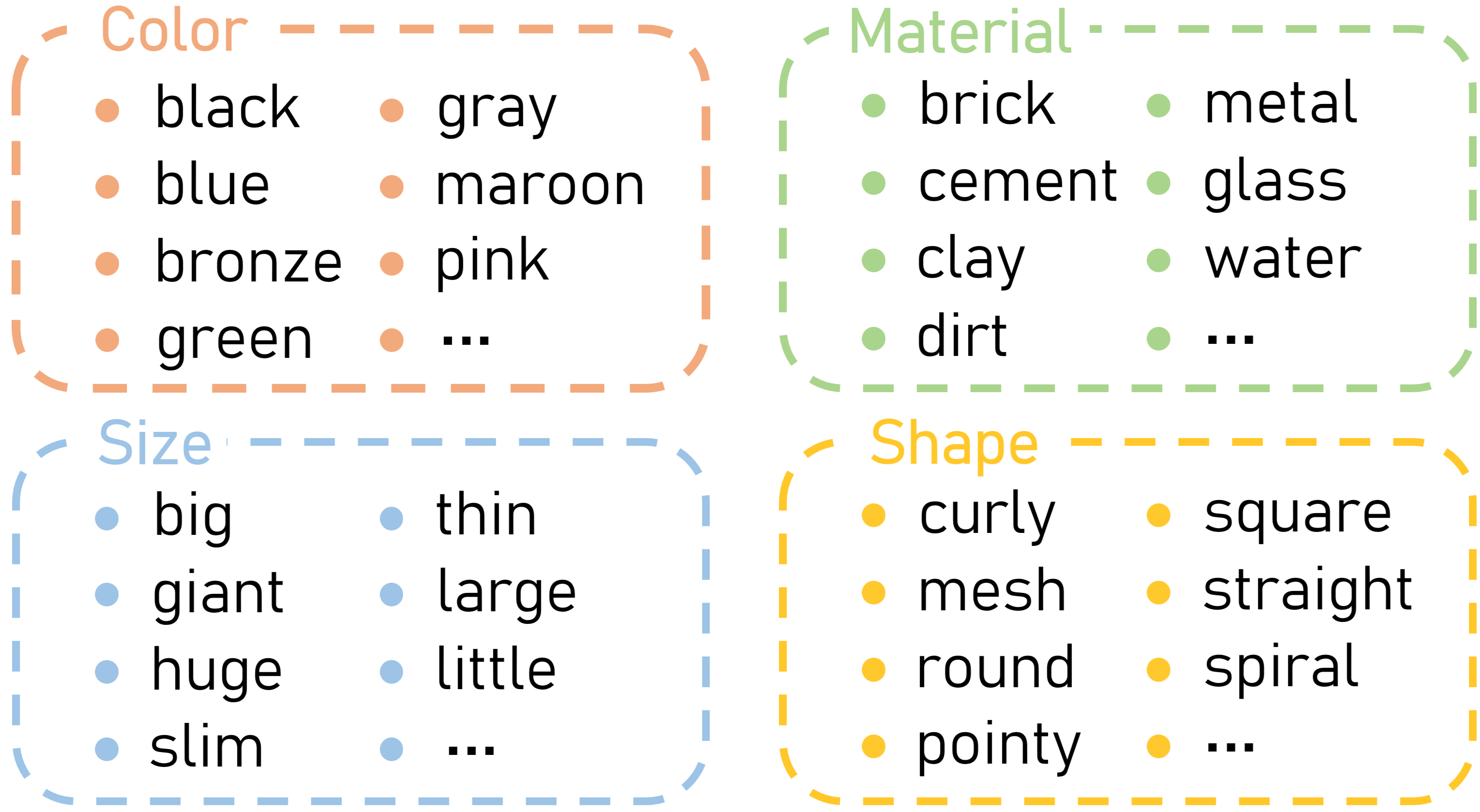}}
\caption{Example text concepts collected from existing visual attribute datasets. Here we present several instances of terms that illustrate \texttt{color}, \texttt{material}, \texttt{size}, and \texttt{shape} within our dictionary of text concepts.}
\label{fig:concept}
\end{center}
\end{figure}

\subsection{Background}
\label{sec:background}
\paragraph{CLIP.} CLIP~\cite{radford2021learning} stands out as a foundational model that constructs an shared embedding space through the fusion of visual and semantic understanding.
This architecture is composed of two encoders: a visual encoder denoted as $E_v$ responsible for handling image input $x$, and a text encoder referred to as $E_t$ designed to process the corresponding textual prompt $t_c$ built as ``\texttt{a photo of} $[\textit{CLS}]_c$", where $[\textit{CLS}]_c$ represents the word embedding for the class $c$. During training, CLIP learns to optimize the resemblance between the image feature and the prompt embeddings associated with the true label.

\paragraph{CoOp and CoCoOp.} CoOp~\cite{zhou2022learning} replaces manual prompt construction by introducing an alternate method that involves learned prompts. This method utilizes a collection of $n$ adaptable context vectors 
$\{[V_{1}], [V_{2}],\cdots , [V_{n}]\}$, each having the same dimension as word embeddings. These vectors are iteratively updated through gradient descent. For a specific class $c$, the respective prompt can be represented as 
$t_{c} = \{[V_{1}], [V_{2}],\cdots , [V_{n}], [\textit{CLS}]_c\}$. CoCoOp~\cite{zhou2022conditional} integrates visual features into prompt learning by utilizing a meta-network $h_{\theta}(x)$ that generates a meta-token $\pi$, denoted as $\pi = h_{\theta}(x)$. The meta-token, combined with the context vectors, form the textual prompts $t_c = \{[V_{1}(x)], [V_{2}(x)], \cdots, [V_n(x)], [\textit{CLS}]_c\}$, where $V_n(x) = V_{n}+\pi$ represents the $n^{th}$ text token.

\subsection{Concept-Guided Prompt Learning}
\label{sec:ADS}
\paragraph{Overview.}
In Figure \ref{fig:overview}, we present an overview of our proposed CPL method.
Figure \ref{fig:overview} (a) shows the visual concept cache establishing process. We first construct a list of text concepts $\Psi_t$ that describe major visual concepts. Then we leverage CLIP's robust text-image correlation capability to discover the image feature $v_j$ with the highest similarity score for each text concept feature $c_t^i \in C_t$. These ``matched'' features are stored in the visual concepts cache as keys, with their corresponding text concepts $\psi_i \in \Psi_t$ as values. Figure \ref{fig:overview} (b) shows the concept-guided discovery process: we first extract the image feature $v$ by $E_v$, then use the image feature as the query to find Top-$K$ similar keys using cosine distance, and finally we utilize the corresponding values to generate a concept-guided prompt. Figure \ref{fig:overview} (c) presents the training pipeline for CPL. We first extract the visual features for a given image $x$ using the visual encoder, then we can obtain the concatenated outputs of different layers $\hat{E}(x)$ as the multi-level features. Next, we follow (b) to generate the concept-guided prompt and extract text features by $E_t$. These features are used as input for the projector, which is a transformer decoder for mapping multi-level visual features into the textual feature space, providing the multi-level visual context. Combined with the multi-level visual context and a task adapter, refined text features work as a classifier for final prediction.

\paragraph{Visual Concept Cache.} 
In Figure \ref{fig:overview} (a), following~\citet{zhang2023cross}, we start by constructing a comprehensive list $\Psi_t$ comprising $I$ text concepts that describe major visual concepts. This list $\Psi_t$ incorporates 2000 common text descriptions for visual concepts gathered from established visual concept datasets~\cite{zhao2019large,pham2021learning}. The descriptions encompass words representing materials, colors, shapes, \etc Illustrations of these terms can be found in Figure \ref{fig:concept}. The dictionary is represented as $\Psi_t \triangleq \{\psi_i\}_{i=1}^I$. Adhering to CLIP's zero-shot setup, we begin by appending $\psi_i$ to a manually designed prompt $\phi =$ ``\texttt{The photo is} ..." to form a concept-specific textual input $\{\phi; \psi_i\}$. Consequently, utilizing the text encoder $E_t$, we generate text concept features $C_t \triangleq \{ c_t^i \}_{i=1}^I$, denoted as $c_t^i = E_t({\phi; \psi_i})$.
 
Within CPL, the visual concepts are discovered by leveraging the text concept features $C_t$ and the CLIP model, derived from the training images. In the scenario of $N$-shot $D$-class few-shot learning, where there exist $N$ labeled images within each of the $D$ classes, the training set is denoted as $T_r\triangleq\{x_j\}_{j=1}^{ND}$. Utilizing the CLIP visual encoder $E_v$, we generate their respective image features $V \triangleq \{ v_j \}_{j=1}^{ND}$, expressed as $v_j = E_v(x_j)$. For every text concept feature $c_t \in C_t$, the similarity score $S_t$ is calculated against all visual features in $V$ using the formula $S_t = \mathrm{sim}\left(c_t,v_j \right)= c_tv_j$, where both $c_t$ and $v_j$ are normalized. Subsequently, we identify the image feature with the highest similarity score as the key and its corresponding text concept word $\psi$ as the associated value, stored within the visual concept cache.

\paragraph{Projector for Vision-to-Language Prompting.}  
Incorporating depictions of rich visual semantics can enhance the precision of the textual content. Multi-level visual features provide richer visual semantics than only high-level (class-specific) features. Therefore, we explore how to utilize multi-level features to optimize the text features. Obviously, we can use a projector to transform multi-level features into the space of text features. Transformer decoder~\cite{vaswani2017attention,rao2022denseclip,lu2021simpler} can model the interactions between vision and language by adopting cross-attention mechanism. Hence, we use the Transformer decoder as the projector. Several studies~\cite{lin2017feature,wang2021pyramid} have already demonstrated that in deep neural networks, the features generated by the earlier layers differ from those produced by the subsequent layers in level. Typically, the earlier layers yield low-level features, such as edges and colors, whereas the later layers produce high-level features, referred to as class-specific features.

Inspired by~\citet{Singha_2023_CVPR}, our aim is to incorporate the multi-level visual features from $E_v$ into the projector $\mathbf{P}$. To achieve this, we utilize global average pooling to reduce the spatial dimensions of individual channels. This produces $\hat{E}_v^{q}(x) \in \mathbb{R}^{C \times 1}$, where $E_v^q\in \mathbb{R}^{W \times H \times C}$ signifies the output derived from the $q^{th}$ layer, where $W$, $H$, $C$ are the width, height, and number of channels of the feature maps. Incorporating this method, we formulate $\hat{E}(x)$ as the concatenation of multi-level features acquired from all $Q$ encoder layers within $E_v$, denoted as $[\hat{E}_v^1(x); \cdots ;\hat{E}_v^Q(x)]$.
We subsequently pass $\hat{E}(x)$ and $\mathbf{f_t}$ through the projector $\mathbf{P}$, which generates multi-level visual context $\mathbf{f_{tv}}$:
\begin{equation}
\label{eq:ftv}
     \mathbf{f_{tv}} = \mathbf{P}(\mathbf{f_t}, \hat{E}(x)),
\end{equation}
where $\mathbf{f_{tv}}$ is the extracted visual contexts, and $\mathbf{f_t}$ is the text feature generated by the CLIP text encoder. 
This implementation fosters the exploration of text features to identify the most relevant visual cues.

\paragraph{Task Adapter.}
As illustrated in Figure \ref{fig:overview}, we append a learnable matrix (\textit{i.e.} task adapter), denoted by $A$, to text features $\mathbf{f_t}$ generated by the text encoder $E_t$. $A$ is task-specific and updated by gradient descent during the training process. In this way, it performs directly on the text-based classifier and explicitly decouples the inherent knowledge of the pre-trained models and the
new knowledge for a target task. Therefore, we can preserve the prior knowledge of concept-guided prompts and assimilate knowledge from new tasks, improving the adaptability of the proposed model.

\subsection{CPL Training and Inference}

\paragraph{CPL Training.} In the training phase, we utilize the supervised contrastive loss, represented as $\mathcal{L}_{ce}$, as the loss function for our approach. This cross-entropy loss guarantees an appropriate alignment between visual and textual feature representations.
Given an image $x$, we first generate its visual features by $E_v(x)$, denoted as $\mathbf{f_v}$,  we follow the concept-guided prompt discovery process to find the concept-guided prompt, denoted as $P_c$, then we can obtain the text features by $E_t(P_c)$, denoted as $\mathbf{f_t}$. According to Equation (\ref{eq:ftv}), we get $\mathbf{f_{tv}}$. Finally, we can calculate the refined text features $\widetilde{\mathbf{f_t}}$ by, 
\begin{equation}
    \widetilde{\mathbf{f_t}} = \mathbf{f_t} + \boldsymbol{\alpha}\mathbf{f_{tv}} + \boldsymbol{\beta}A ,
    \label{eq:text}
\end{equation}
where $\boldsymbol{\alpha}$ and $\boldsymbol{\beta}$ are learnable parameters to control the scaling of the residual and text features are updated through a residual connection. The values assigned to parameters $\boldsymbol{\alpha}$ and $\boldsymbol{\beta}$ upon initialization are exceptionally diminutive (\textit{e.g.}, $10^{-4}$). This choice aims to uphold the language priors extensively within the original text features. The prediction probability for $x$ pertaining to label $i$ is represented as
\begin{equation}
p(y=i|x) = \frac{ \exp(\text{sim}(\widetilde{\mathbf{f_t^i}}, \mathbf{f_v})/\tau)}{\sum_{j=1}^{K}\ \exp(\text{sim}(\widetilde{\mathbf{f_t^j}}, \mathbf{f_v})/ \tau))},
\label{eq:prob}
\end{equation}
where $\tau$ is a temperature coefficient and `$\text{sim}$' represents the cosine similarity.  The cross-entropy loss is calculated by
\begin{equation}
    \mathcal{L}_{ce} = - \underset{\theta_\mathbf{P},\ A}{\arg\min} \  \underset{(x,y) \in \mathcal{D}_{tr}}{\mathbb{E}} \sum_{k=1}^{\mathcal{Y}_{tr}} y_{k} \log(p(y_k|x)),
    \label{eq:lce}
\end{equation}
where $\theta_\mathbf{P}$ is the parameter weights of the projector $\mathbf{P}$, $A$ is the learnable matrix of the task adapter, and $\mathcal{Y}_{tr}$ are the class labels for the training dataset.

\paragraph{CPL Inference.} During the inference phase for the test dataset $\mathcal{D}_{te}$, where $\mathcal{Y}_{te}$ signifies the labels of this dataset, we calculate the cosine similarity between the images $x_{te}$ and the prompt embeddings for all classes within the test dataset $\mathcal{Y}_{te}$. The class exhibiting a higher probability value is subsequently chosen:
\begin{equation}
    \hat{y}_{te} = \underset{y \in \mathcal{Y}_{te}} {\arg\max} \ p(y|x_{te}).
    \label{yte}
\end{equation}

\section{Experiments}
\subsection{Benchmark Settings}
\paragraph{Task Settings.} We follow previous work to evaluate our proposed approach on four challenging task settings:
\begin{itemize}
    \item \textbf{Generalization from Base to Novel Classes.} We evaluate the generalization capability of our method in a zero-shot scenario by dividing the datasets into base and novel classes. We train our model with few-shot images on the base classes and then evaluate on unseen novel classes.
    \item \textbf{Cross-Dataset Transfer.} We conduct a direct evaluation of our ImageNet-trained model across various other datasets. Following previous work, we train our model on all 1,000 ImageNet classes in a few-shot setting.
    \item \textbf{Domain Generalization.} We evaluate the robustness of our method on OOD datasets. Similarly, we evaluate our model trained on ImageNet directly on four ImageNet variants that encompass different types of shifts.
\end{itemize}

\paragraph{Datasets.} For base-to-novel generalization, cross-dataset transfer tasks, we follow previous work~\cite{radford2021learning,zhou2022learning,zhou2022conditional} to conduct the experiments on 11 representative image classification datasets, including ImageNet~\cite{deng2009imagenet} and Caltech101~\cite{fei2004learning} for generic object classification; OxfordPets~\cite{parkhi2012cats}, StanfordCars~\cite{krause20133d}, Flowers102~\cite{nilsback2008automated}, Food101~\cite{bossard2014food}, and FGVCAircraft~\cite{maji2013fine} for fine-grained classification; SUN397~\cite{xiao2010sun} for scene recognition; UCF101~\cite{soomro2012ucf101} for action recognition; DTD~\cite{cimpoi2014describing} for texture classification; and EuroSAT~\cite{helber2019eurosat} for satellite image recognition. For domain generalization, we utilize ImageNet as the source dataset and four ImageNet variants as target datasets including ImageNet-A~\cite{hendrycks2021natural},  ImageNet-R~\cite{hendrycks2021many}, ImageNet-V2~\cite{recht2019imagenet}, ImageNet-Sketch~\cite{wang2019learning}. 

\begin{table*}[t!]
\centering
\renewcommand{\arraystretch}{0.82}
{\fontsize{9pt}{11pt}\selectfont
    
    \begin{subtable}[t]{.32\textwidth}
    \centering
    \caption{\textbf{Average over 11 datasets.}}
    \begin{tabular}{l cc|c}
    \toprule
    Method & Base & Novel & HM \\
    \midrule
    CLIP & 69.34 & 74.22 & 71.70 \\
    CoOp & \underline{82.69} & 63.22 & 71.66 \\
    CoCoOp & 80.47 & 71.69 & 75.83 \\
    MaPLe & 82.28 & \underline{75.14} & \underline{78.55} \\
    ProGrad & 82.48 & 70.75 & 76.16 \\
    KgCoOp & 80.73 & 73.60 & 77.00 \\
    \midrule
    
    Ours & \textbf{84.38} & \textbf{78.03} & \textbf{81.08} \\
    &  \text{{+1.69}} &  \text{{+2.89}} &  \text{{+2.53}} \\
    \bottomrule
    \end{tabular}
    \end{subtable}
    \begin{subtable}[t]{.32\textwidth}
    \centering
    \caption{ImageNet.}
    \begin{tabular}{l cc|c}
    \toprule
    Method & Base & Novel & HM \\
    \midrule
    CLIP & 72.43 & 68.14 & 70.22 \\
    CoOp & {76.47} & 67.88 & 71.92\\
    CoCoOp & 75.98 & {70.43} & {73.10} \\
    MaPLe & {76.66} & \underline{70.54} & \underline{73.47} \\
    ProGrad & \underline{77.02} & 66.66 & 71.46 \\
    KgCoOp & 75.83 & 69.96 & 72.78 \\
    \midrule
    
    Ours & \textbf{78.74} & \textbf{72.03} & \textbf{75.24} \\
      &  \text{{+1.72}} &  \text{{+1.49}} &  \text{{+1.77}} \\
    \bottomrule
    \end{tabular}
    \end{subtable}
    ~
    \begin{subtable}[t]{.32\textwidth}
    \centering
    \caption{Caltech101.}
    \begin{tabular}{l cc|c}
    \toprule
    Method & Base & Novel & HM \\
    \midrule
    CLIP & 96.84 & {94.00} & 95.40 \\
    CoOp & {98.00} & 89.81 & 93.73 \\
    CoCoOp & 97.96 & 93.81 & {95.84} \\
    MaPLe & 97.74 & {94.36} & {96.02} \\
    ProGrad & \underline{98.02} & 93.89 & 95.91 \\
    KgCoOp & 97.72 & \underline{94.39} & \underline{96.03} \\
    \midrule
    
    Ours &  \textbf{98.35} & \textbf{95.13} & \textbf{96.71} \\
      &  \text{{+0.33}} &  \text{{+0.74}} &  \text{{+0.68}} \\
    \bottomrule
    \end{tabular}
    \end{subtable}
    ~
    \begin{subtable}[t]{.32\textwidth}
    \centering
    \caption{OxfordPets.}
    \begin{tabular}{l cc|c}
    \toprule
    Method & Base & Novel & HM \\
    \midrule
    CLIP & 91.17 & 97.26 & 94.12 \\
    CoOp & 93.67 & 95.29 & 94.47 \\
    CoCoOp & {95.20} & {97.69} & {96.43} \\ 
    MaPLe & \underline{95.43} & {97.76} & \underline{96.58}\\
    ProGrad & 95.07 & 97.63 & 96.33 \\
    KgCoOp & 94.65 & \underline{97.76} & 96.18 \\
    \midrule
        
    Ours & \textbf{95.86} & \textbf{98.21} & \textbf{97.02} \\
     &  \text{{+0.43}} &  \text{{+0.45}} &  \text{{+0.44}} \\
    \bottomrule
    \end{tabular}
    \end{subtable}
    \begin{subtable}[t]{.32\textwidth}
    \centering
    \caption{StanfordCars.}
    \begin{tabular}{l cc|c}
    \toprule
    Method & Base & Novel & HM \\
    \midrule
    CLIP & 63.37 & {74.89} & 68.65 \\
    CoOp & \underline{78.12} & 60.40 & 68.13 \\
    CoCoOp & 70.49 & 73.59 & {72.01} \\
    MaPLe & 72.94 & 74.00 & \underline{73.47} \\
    ProGrad & 77.68 & 68.63 & 72.88 \\
    KgCoOp & 71.76 & \underline{75.04} & 73.36 \\
    \midrule
        
    Ours & \textbf{79.31} & \textbf{76.65} & \textbf{77.96} \\
    &  \text{{+1.19}} &  \text{{+1.61}} &  \text{{+4.49}} \\
    \bottomrule
    \end{tabular}
    \end{subtable}
    ~
    \begin{subtable}[t]{.32\textwidth}
    \centering
    
    \caption{Flowers102.}
    
    \begin{tabular}{l cc|c}
    \toprule
    Method & Base & Novel & HM \\
    \midrule
    CLIP & 72.08 & \underline{77.80} & 74.83 \\
    CoOp & \underline{97.60} & 59.67 & 74.06 \\
    CoCoOp & 94.87 & 71.75 & {81.71} \\
    MaPLe & 95.92 & 72.46 & {82.56} \\
    ProGrad & 95.54 & 71.87 & 82.03 \\
    KgCoOp & 95.00 & 74.73 & \underline{83.65} \\
    \midrule
        
    Ours & \textbf{98.07} & \textbf{80.43} & \textbf{88.38} \\
     &  \text{{+0.47}} &  \text{{+2.63}} &  \text{{+4.73}} \\
    \bottomrule
    \end{tabular}
    \end{subtable}
    ~
    \begin{subtable}[t]{.32\textwidth}
    \centering
    
    \caption{Food101.}
    
    \begin{tabular}{l cc|c}
    \toprule
    Method & Base & Novel & HM \\
    \midrule
    CLIP & 90.10 & 91.22 & 90.66 \\
    CoOp & 88.33 & 82.26 & 85.19 \\
    CoCoOp & {90.70} & {91.29} & {90.99} \\
    MaPLe & \underline{90.71} & \underline{92.05} & \underline{91.38} \\
    ProGrad & 90.37 & 89.59 & 89.98 \\
    KgCoOp & 90.05 & 91.70 & 91.09 \\
    \midrule
        
    Ours & \textbf{91.92} & \textbf{93.87} & \textbf{92.88} \\
     &  \text{{+1.21}} &  \text{{+1.82}} &  \text{{+1.50}} \\
    \bottomrule
    \end{tabular}
    \end{subtable}
    \begin{subtable}[t]{.32\textwidth}
    \centering
    
    \caption{FGVCAircraft.}
    
    \begin{tabular}{l cc|c}
    \toprule
    Method & Base & Novel & HM \\
    \midrule
    CLIP & 27.19 & \underline{36.29} & {31.09} \\
    CoOp & {40.44} & 22.30 & 28.75 \\
    CoCoOp & 33.41 & 23.71 & 27.74 \\
    MaPLe & 37.44 & 35.61 & \underline{36.50} \\
    ProGrad & \underline{40.54} & 27.57 & 32.82 \\
    KgCoOp & 36.21 & 33.55 & 34.83 \\
    \midrule
        
    Ours & \textbf{42.27} & \textbf{38.85} & \textbf{40.49} \\
      &  \text{{+1.73}} &  \text{{+2.56}} &  \text{{+3.99}} \\
    \bottomrule
    \end{tabular}
    \end{subtable}
    ~
    \begin{subtable}[t]{.32\textwidth}
    \centering
    
    \caption{DTD.}
    
    \begin{tabular}{l cc|c}
    \toprule
    Method & Base & Novel & HM \\
    \midrule
    CLIP & 53.24 & \underline{59.90} & 56.37 \\
    CoOp & {79.44} & 41.18 & 54.24 \\
    CoCoOp & 77.01 & 56.00 & {64.85} \\
    MaPLe & \underline{80.36} & 59.18 & \underline{68.16} \\
    ProGrad & 77.35 & 52.35 & 62.45 \\
    KgCoOp & 77.55 & 54.99 & 64.35 \\
    \midrule
        
    Ours & \textbf{80.92} & \textbf{62.27} & \textbf{70.38} \\
      &  \text{{+0.56}} &  \text{{+2.37}} &  \text{{+2.22}} \\
    \bottomrule
    \end{tabular}
    \end{subtable}
    ~
    \begin{subtable}[t]{.32\textwidth}
    \centering
    
    \caption{SUN397.}
    
    \begin{tabular}{l cc|c}
    \toprule
    Method & Base & Novel & HM \\
    \midrule
    CLIP & 69.36 & 75.35 & 72.23 \\
    CoOp & {80.60} & 65.89 & 72.51 \\
    CoCoOp & 79.74 & {76.86} & {78.27} \\
    MaPLe & 80.82 & \underline{78.70} & \underline{79.75} \\
    ProGrad & \underline{81.26} & 74.17 & 77.55 \\
    KgCoOp & 80.29 & 76.53 & 78.36\\
    \midrule
        
    Ours & \textbf{81.88} & \textbf{79.65} & \textbf{80.75} \\
      &  \text{{+0.62}} &  \text{{+0.95}} &  \text{{+1.00}} \\
    \bottomrule
    \end{tabular}
    \end{subtable}
    ~
    \begin{subtable}[t]{.32\textwidth}
    \centering
    
    \caption{EuroSAT.}
    
    \begin{tabular}{l cc|c}
    \toprule
    Method & Base & Novel & HM \\
    \midrule
    CLIP & 56.48 & {64.05} & 60.03 \\
    CoOp & {92.19} & 54.74 & 68.69 \\
    CoCoOp & 87.49 & 60.04 & {71.21} \\
    MaPLe & \underline{94.07} & \underline{73.23} & \underline{82.35} \\
    ProGrad & 90.11 & 60.89 & 72.67 \\
    KgCoOp & 85.64 & 64.34 & 73.48 \\
    \midrule
        
    Ours & \textbf{94.18} & \textbf{81.05} & \textbf{87.12} \\
      &  \text{{+0.11}} &  \text{{+7.82}} &  \text{{+4.77}} \\
    \bottomrule
    \end{tabular}
    \end{subtable}
    ~
    \begin{subtable}[t]{.32\textwidth}
    \centering
    
    \caption{UCF101.}
    
    \begin{tabular}{l cc|c}
    \toprule
    Method & Base & Novel & HM \\
    \midrule
    CLIP & 70.53 & {77.50} & 73.85 \\
    CoOp & \underline{84.69} & 56.05 & 67.46 \\
    CoCoOp & 82.33 & 73.45 & {77.64} \\
    MaPLe & 83.00 & \underline{78.66} & \underline{80.77} \\
    ProGrad & 84.33 & 74.94 & 79.35 \\
    KgCoOp & 82.89 & 76.67 & 79.65 \\
    \midrule
        
    Ours & \textbf{86.73} & \textbf{80.17} & \textbf{83.32} \\
     &  \text{{+2.04}} &  \text{{+1.51}} &  \text{{+2.55}} \\
    \bottomrule
    \end{tabular}
    \end{subtable}
    }
    \caption{Comparison with state-of-the-art methods on base-to-novel generalization (on ViT-B/16 backbone). Our proposed method learns local concepts and demonstrates strong generalization results over existing methods on 11 recognition datasets. The best results are in bold and the second-best results are underlined.}
    \label{tab:base2novel}
\end{table*}

\paragraph{Implementation Details.}
For a fair comparison, we use the ViT-B/16 CLIP model for base-to-novel generalization and cross-dataset transfer and the ResNet-50 CLIP model for domain generalization.
Throughout the training process, both the visual and textual encoders remain fixed.
We adhere to the data pre-processing protocol outlined in CLIP, which involves resizing and applying random cropping operations, \etc.  We conduct training for 70 epochs on the ImageNet and 50 epochs for other datasets. 
We designate the number of concepts $K$ as 10. Training involves a batch size of 256 and an initial learning rate set at $10^{-3}$. We employ the AdamW optimizer with a cosine annealing scheduler and train the models on a single NVIDIA RTX 3090 GPU. 
Code will be available at https://github.com/rambo-coder/CPL.

\subsection{Generalization from Base to Novel Classes}\

\begin{table*}[!h]
    \resizebox{\linewidth}{!}{
    \begin{tabular}{l c ccccccccccc}
    \toprule
    & \textbf{Source} & \multicolumn{11}{c}{\textbf{Target}} \\ \cmidrule(lr){2-2} \cmidrule(lr){3-13}
    & \rotbox{ImageNet} & \rotbox{Caltech101} & \rotbox{OxfordPets} & \rotbox{StanfordCars} & \rotbox{Flowers102} & \rotbox{Food101} & \rotbox{Aircraft} & \rotbox{SUN397} & \rotbox{DTD} & \rotbox{EuroSAT} & \rotbox{UCF101} & \rotbox{\emph{Average}} \\
    \midrule
    CoOp & \underline{71.51} & 93.70 & 89.14 & 64.51 & 68.71 & 85.30 & 18.47 & 64.15 & 41.92 & {46.39} & 66.55 & 63.88 \\
    CoCoOp & 71.02 &\underline{94.43} & 90.14 & 65.32 & 71.88 & 86.06 & 22.94 & \underline{67.36} & 45.73 & 45.37 & 68.21 & {65.74} \\
    MaPLe & 70.72 & 93.53 & \underline{90.49} & \underline{65.57} & \underline{72.23} & \underline{86.20} & \underline{24.74} & {67.01} & \underline{46.49} & \underline{48.06} & \underline{68.69} & \underline{66.30} \\
 \textbf{Ours} & \textbf{73.53} & \textbf{95.52} & \textbf{91.64} & \textbf{66.17} & \textbf{73.35} & \textbf{87.68} & \textbf{27.36} & \textbf{68.24} & \textbf{48.96} & \textbf{51.25} & \textbf{70.52} & \textbf{68.07} \\
    \bottomrule
    \end{tabular}}
        \caption{Comparison of our method with existing approaches on cross-dataset evaluation. Overall, our method demonstrates superior generalization capabilities with the highest average accuracy on 10 datasets.}
    \label{tab:crossdata}
\end{table*}

We compare our method with six baselines: zero-shot CLIP~\cite{radford2021learning}, CoOp~\cite{zhou2022learning}, CoCoOp~\cite{zhou2022conditional}, ProGrad~\cite{zhu2023prompt}, MaPLe~\cite{khattak2023maple}, and KgCoOp~\cite{yao2023visual}. Table \ref{tab:base2novel} displays results regarding base-to-novel generalization across 11 datasets with 16-shot samples.

\paragraph{Performance Evaluation on Base Classes.} CoOp demonstrates remarkable performance on base classes among previous methods. However, it exhibits an overfitting problem for its excessive dependence on a single learnable prompt component, as argued by CoCoOp. Our method remarkably surpasses CoOp by an average accuracy gain of 1.69\% without a generalizability depletion, and achieves the best performance on base classes for all datasets, as illustrated in Table \ref{tab:base2novel}. Our method's efficacy indicates its substantial capability to adapt to downstream tasks. 

\paragraph{Generalization to Unseen Classes.} Although CoCoOp improves CoOp's limited generalizability by conditioning prompts on image instances, it has an average degradation of -2.22\% on base classes. As a comparison, MaPLe obtains balanced performance on both base and novel classes. Remarkably, our CPL method achieves the highest performance in terms of novel classes and harmonic mean (\textbf{HM}) on all 11 datasets, with an accuracy improvement of 2.89\% and 2.53\%, respectively. With visual concepts extracted from prior knowledge, our CPL can better generalize to novel categories. The exceptional performance demonstrates the enhanced generalizability of CPL to unseen classes without sacrificing performance in base classes.

\begin{table}[t]
\begin{center}

\resizebox{\linewidth}{!}{
\begin{tabular}{lccccc}
\toprule
\multirow{2}*{Method}  & Source & \multicolumn{4}{c}{Target} \\ \cmidrule(lr){2-2} \cmidrule(lr){3-6} & ImageNet & -V2 & -Sketch & -A & -R  \\
\midrule
CLIP &  60.33 &  53.27  & \underline{35.44}  & 21.65 &  56.00  \\
CoOp   & 63.33  & 55.40  & 34.67  & 23.06  & 56.60 \\
CoCoOp  & 62.81  & 55.72  & 34.48  & 23.32  & 57.74  \\
ProGrad   & 62.17  & 54.70  & 34.40  & 23.05  & 56.77  \\
PLOT   & 63.01  & 55.11  & 33.00  & 21.86  & 55.61  \\
DeFo    & \underline{64.00}  & \underline{58.41}  & 33.18  & 21.68  & 55.84 \\
TPT  & 60.74  & 54.70  & 35.09  & \underline{26.67}  & \underline{59.11} \\


\textbf{Ours}  & \textbf{66.92}  & \textbf{58.67}  & \textbf{37.64}  & \textbf{31.05}  & \textbf{60.08}\\
\bottomrule

\end{tabular}
}
\end{center}
\caption{Comparison with other methods on robustness ($\%$) to natural distribution shifts. The best results are in bold and the second-best results are underlined.}
\label{table:generalization}
\end{table}

\begin{table}[t]

\centering
\resizebox{\linewidth}{!}{
\begin{tabular}{lccccc}
\toprule
Method & 1    & 2  & 4  & 8 & 16 \\ \midrule
CLIP                    & 60.33 & 60.33 & 60.33 & 60.33 & 60.33 \\
\, + CGP               & 61.06 & 61.59 & 62.65 & 63.17 & 64.38 \\
\, + CGP + P          & 62.32   & 62.88     & 63.80   & 64.83    & 66.35    \\

\, + CGP + P + TA    & \textbf{63.02} & \textbf{63.37} & \textbf{64.36} & \textbf{65.31} & \textbf{66.92} \\
\bottomrule
\end{tabular}
}
\caption{Effectiveness of different components in our method. CGP and P represent concept-guided prompting and the projector, respectively, and TA is the task adapter.}
\label{table:components}
\end{table}

\subsection{Cross-Dataset Transfer}
Cross-dataset transfer is a much more challenging generalization task compared to base-to-novel generalization, since the latter only transfers within a single dataset while the former transfers across different datasets, \eg, from object recognition to scene classification. We test the cross-dataset generalization ability of our method on 1000 ImageNet classes and then transfer it directly to the remaining 10 datasets. The comparison results with CoOp, CocoOp, and MaPLe are presented in Table \ref{tab:crossdata}. Overall, our CPL method marks the best performance on both source and target datasets with a target average of 68.07\%, and outperforms MaPLe by 1.77\%. Notably, our method surpasses MaPLe by 3.2\% on EuroSAT, a satellite image dataset whose fundamentals are distinctive from ImageNet. This suggests that concept-guided prompting in our method facilitates better generalization, as illustrated in Figure \ref{fig:comparison}.


\subsection{Domain Generalization}

In Table \ref{table:generalization}, we provide the classification accuracy across the source domain and target domains, as well as the average accuracy within target domains (OOD Average).
In addition to the methods mentioned earlier, we also compare our approach with PLOT~\cite{chen2023plot}, DeFo~\cite{wang2023learning}, TPT~\cite{shu2022test}. Our approach surpasses other methods in all scenarios, indicating the remarkable robustness of our model against distribution shifts.

\subsection{Ablation Studies}

\paragraph{Contributions of major algorithm components. } From Table \ref{table:components}, we can see that all three components contribute significantly to the enhanced performance. Among them, concept-guided prompting brings the largest performance improvement, for example, a 4.05\% improvement in 16-shot accuracy. This shows that a more accurate and specific text description leads to better classification results.


\paragraph{The number $K$ of concepts selected and the size $I$ of text concepts set. } We investigate the impact of $K$ by varying the number of concepts selected and show the results in Table \ref{table:conceptnum}. We find that our method achieves the best performance when $K = 10$. The results also show that our method achieves the best performance when $I = 2000$. When the size is too large, the performance decreases since the different text concepts might match the same visual concept.

\begin{table}[t]
\centering
\begin{tabular}{c|ccccc}
\toprule
\textbf{Value of $K$}  & 6   & 8 & \textbf{10}  & 12 & 14 \\ \midrule
\textbf{Accuracy}  & 65.33 & 66.28 & \textbf{66.92} & 66.56 & 66.31 \\ \midrule
\textbf{Value of $I$}  & 1000   & 1500 & \textbf{2000} & 2500 & 3000 \\ \midrule
\textbf{Accuracy}  & 63.67 & 65.88 & \textbf{66.92} & 66.71 & 66.23 \\
\bottomrule
\end{tabular}
\caption{
Number $K$ of concepts selected and total size $I$ of concept set. Experiments are conducted on 16-shot ImageNet. Here for $I$, we fix the number of concept categories and vary the number of concepts in each category.}
\label{table:conceptnum}
\end{table}

\paragraph{Comparison on the number of training epochs and time.} As shown in Table \ref{tab:efficiency}, our proposed CPL outperforms other methods by a large margin with only 50 minutes, while CoOp and ProGrad need more than 14 hours. This demonstrates the remarkable efficiency of our method.

\begin{table}[t]
\centering

\resizebox{\linewidth}{!}{
\begin{tabular}{lccccccc}
\toprule 
Method  &  Epochs  &  Time  &  Accuracy  &  Gain \\
\midrule
Zero-shot CLIP  &  0  &  0  &  60.33  &  0 \\
Linear Probe CLIP  &   -  &  13m  &  56.13  &  -4.20 \\
CoOp  &   200  &  14h 40m  &  62.26   &  +1.93 \\
ProGrad  &   200  &  17h  &  63.45   &  +3.12 \\

\textbf{Ours}  &   70  &  50min  &  \textbf{66.92}  &  \textbf{+6.59} \\
    \bottomrule
    \end{tabular}
}
\caption{Comparison on the number of training epochs and time on 16-shot ImageNet. }
\label{tab:efficiency}

\end{table}

\section{Conclusion}
In this work, we introduce Concept-Guided Prompt Learning (CPL) for vision-language models. By utilizing the profound knowledge embedded in CLIP, we form a visual concept cache that facilitates concept-guided prompting. To further refine text features, we design a projector that projects multi-level visual features into corresponding textual features.
Our proposed CPL method exhibits great effectiveness in diverse applications such as base-to-novel generalization, cross-dataset transfer, and domain generalization tasks. 
Supported by thorough experimental analysis, we demonstrate that our proposed CPL achieves remarkable performance improvements, and also surpasses existing leading-edge methods by substantial margins. 
 
\bibliography{aaai24}
\end{document}